\documentclass[11pt]{article} 
\usepackage{bvsr_arxiv}

\usepackage{amsmath,amsfonts,bm}

\def\eqref#1{equation~\ref{#1}}
\def\Eqref#1{Equation~\ref{#1}}

\def\1{\bm{1}}

\DeclareMathAlphabet{\mathsfit}{\encodingdefault}{\sfdefault}{m}{sl}
\SetMathAlphabet{\mathsfit}{bold}{\encodingdefault}{\sfdefault}{bx}{n}

\usepackage{graphicx,tabularray}
\UseTblrLibrary{booktabs}
\usepackage{url}
\usepackage[colorlinks=true,breaklinks=true]{hyperref}
\hypersetup{allcolors=paperlink}
\usepackage{xcolor}
\usepackage{fancyhdr}
\usepackage[most]{tcolorbox}
\usepackage{tabularx,booktabs}
\definecolor{mblueL}{HTML}{C8D6D7}\definecolor{mblueD}{HTML}{87A3A7}
\definecolor{mtext}{HTML}{30343A}
\definecolor{MyTitleBackGround}{HTML}{efdcf9}
\usepackage{etoc}
\etocsettagdepth{mainpaper}{none}
\etocsettagdepth{appendix}{subsubsection}
\usepackage{graphicx}
\usepackage{booktabs}
\usepackage{caption}
\usepackage{amsthm} 
\newtheorem{theorem}{Theorem}[section]
\newtheorem{proposition}{Proposition}[section]
\usepackage[svgnames]{xcolor} 
\usepackage{amsthm}           
\usepackage[most]{tcolorbox}  
\definecolor{propBack}{HTML}{F4F6F9}
\definecolor{propBar}{HTML}{5E6E82}
\tcbset{stanfordbox/.style={
    enhanced, breakable,
    frame hidden, boxrule=0pt, sharp corners,
    colback=propBack,
    borderline west={2pt}{0pt}{propBar},
    left=8pt, right=8pt, top=5pt, bottom=5pt,
    before skip=10pt plus 2pt,
    after skip=10pt plus 2pt,
}}
\tcolorboxenvironment{theorem}{stanfordbox}
\tcolorboxenvironment{proposition}{stanfordbox}
\newtheorem{assumption}{Assumption}[section]
\usepackage{capt-of}
\newsavebox{\ablationfigurebox}
\theoremstyle{remark}
\newtheorem{remark}{Remark}

\title{Blind Variation, Selective Retention:\\Improving Creative Idea Generation in AI Scientists}

\definecolor{figBlue}{HTML}{1C4C8C}
\definecolor{figOrange}{HTML}{984C08}
\definecolor{titleLinkColor}{HTML}{5E6E82}
\newcommand{\githubicon}{\raisebox{-0.1ex}{\small\faGithub}}
\newcommand{\websiteicon}{\raisebox{-0.1ex}{\small\faGlobe}}
\newcommand{\hficon}{\raisebox{-0.22em}[0pt][0pt]{\includegraphics[height=1.1em]{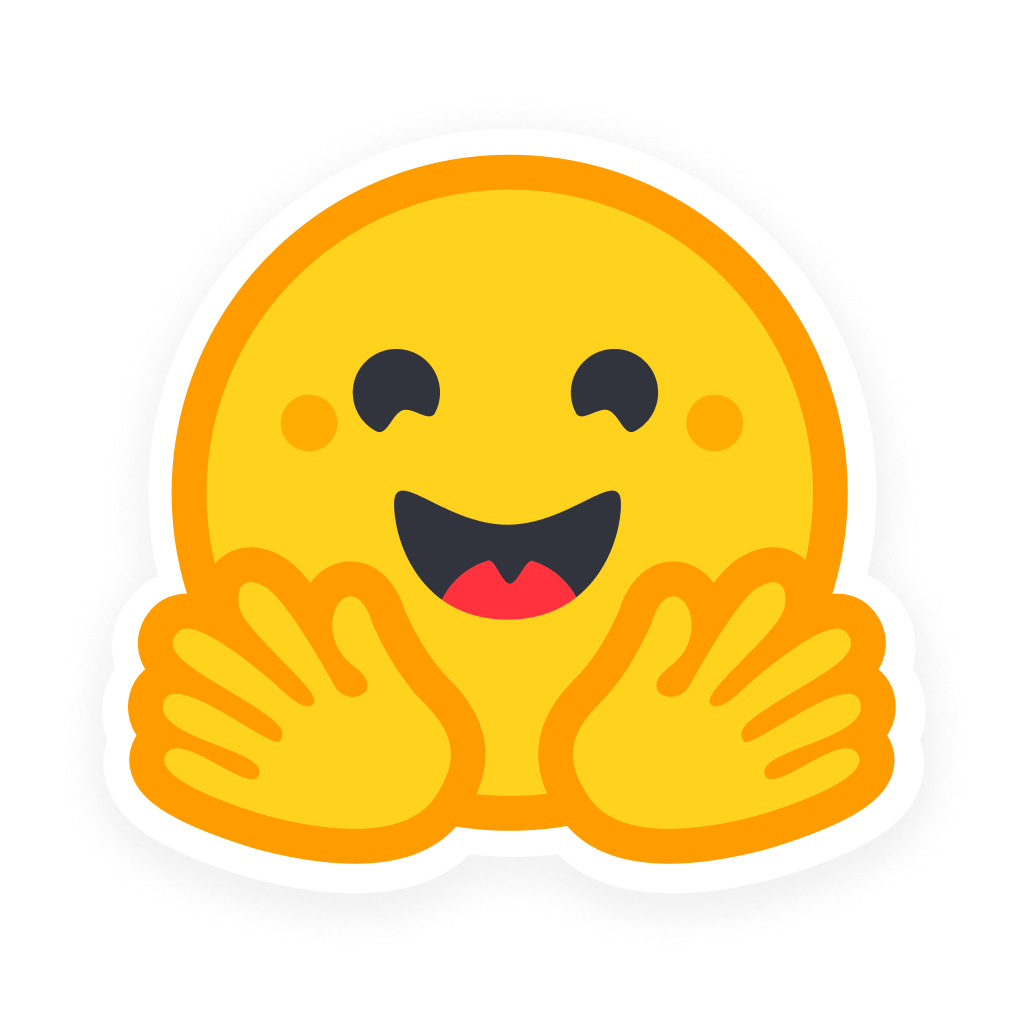}}}

\newcommand{\pullup}[1]{\par\vspace{-#1}\nointerlineskip\hbox{}}

\author{%
  Guowei Xu$^{1,*}$\hspace{1.2em}%
  Jiaru Zou$^{1,*}$\hspace{1.2em}%
  Mete Erdogan$^{1}$\hspace{1.2em}Haotian Ye$^{1}$\hspace{1.2em}Pan Lu$^{1}$\hspace{1.2em}James Zou$^{1}$\\[0.3em]
  $^{1}$Stanford University\qquad $^{*}$Co-first Authors
}

\usepackage{graphicx}
\usepackage{booktabs}
\usepackage{amsmath,amssymb}
\graphicspath{{figures/}}

\definecolor{promptBack}{HTML}{FAFBFC}
\definecolor{promptFrame}{HTML}{C5CDD8}
\definecolor{promptTitleBack}{HTML}{E9EDF2}
\newtcblisting[auto counter,number within=section]{promptbox}[4][mblueL]{%
  listing only, breakable, enhanced,
  colback=promptBack, colframe=promptFrame,
  colbacktitle=promptTitleBack, coltitle=propBar,
  boxrule=0.6pt, titlerule=0.6pt, arc=2pt,
  left=6pt, right=6pt, top=4pt, bottom=4pt, toptitle=2pt, bottomtitle=2pt,
  fonttitle=\small\bfseries, title={Prompt~\thetcbcounter: #3}, label={#4},
  listing options={basicstyle=\ttfamily\scriptsize\color{mtext}, breaklines=true, breakatwhitespace=false,
                   columns=fullflexible, keepspaces=true, showstringspaces=false,
                   breakindent=0pt, postbreak={}, lineskip=-0.5pt}}

\begin{document}

\maketitle
\thispagestyle{fancy}
\fancyhead{}
\renewcommand{\headrulewidth}{0pt}
\setlength{\headheight}{12pt}
\setlength{\headsep}{3mm}
\etocdepthtag.toc{mainpaper}

\vspace{4pt}
\centerline{\normalfont\normalsize\hypersetup{urlcolor=titleLinkColor}%
  \href{https://zou-group.github.io/BVSR/}{\websiteicon\hspace{0.4em}Website}\qquad
  \href{https://github.com/zou-group/BVSR}{\githubicon\hspace{0.4em}GitHub}\qquad
  \href{https://huggingface.co/collections/Xkev/bvsr}{\hficon\hspace{0.3em}Hugging Face}}
\pullup{8pt}
\begin{abstract}
\vspace{-1em}
Scientific discovery requires ideas that are original, useful, and non-obvious. Yet AI scientists are typically given as much task information as possible during idea generation, which can anchor generation toward familiar approaches. Inspired by the cognitive framework of \textcolor{figBlue}{\textbf{Blind Variation}} and \textcolor{figOrange}{\textbf{Selective Retention}} (BVSR), we propose a discovery framework that separates these two roles of information: it generates proposals under deliberately limited task information to broaden exploration, then learns from proposals previously verified on the same task and their outcomes to retain promising candidates for costly empirical evaluation.
We evaluate BVSR across five proposer models and 12 scientific tasks. Limiting task information substantially increases proposal originality and non-obviousness, and produces a larger share of highly creative and high-utility proposals than existing state-of-the-art prompting-based baselines. Learning from prior proposal outcomes on the task further improves candidate selection, discarding half of the candidates while retaining
most top-performing proposals. When incorporated into existing AI-scientist workflows, BVSR achieves stronger discoveries across gene expression analysis, drug response prediction, anomaly detection, scaling law discovery, and brain activity prediction. 
 These results suggest that effective AI scientists should use information differently across stages of discovery: constrain task-specific information during exploration while exploiting accumulated empirical experience on the task.
\end{abstract}
\vspace{-4pt}
\begingroup\setlength{\intextsep}{0pt}%
\begin{figure}[H]
    \captionsetup{skip=15pt}
    \centering    \includegraphics[width=0.98\linewidth]{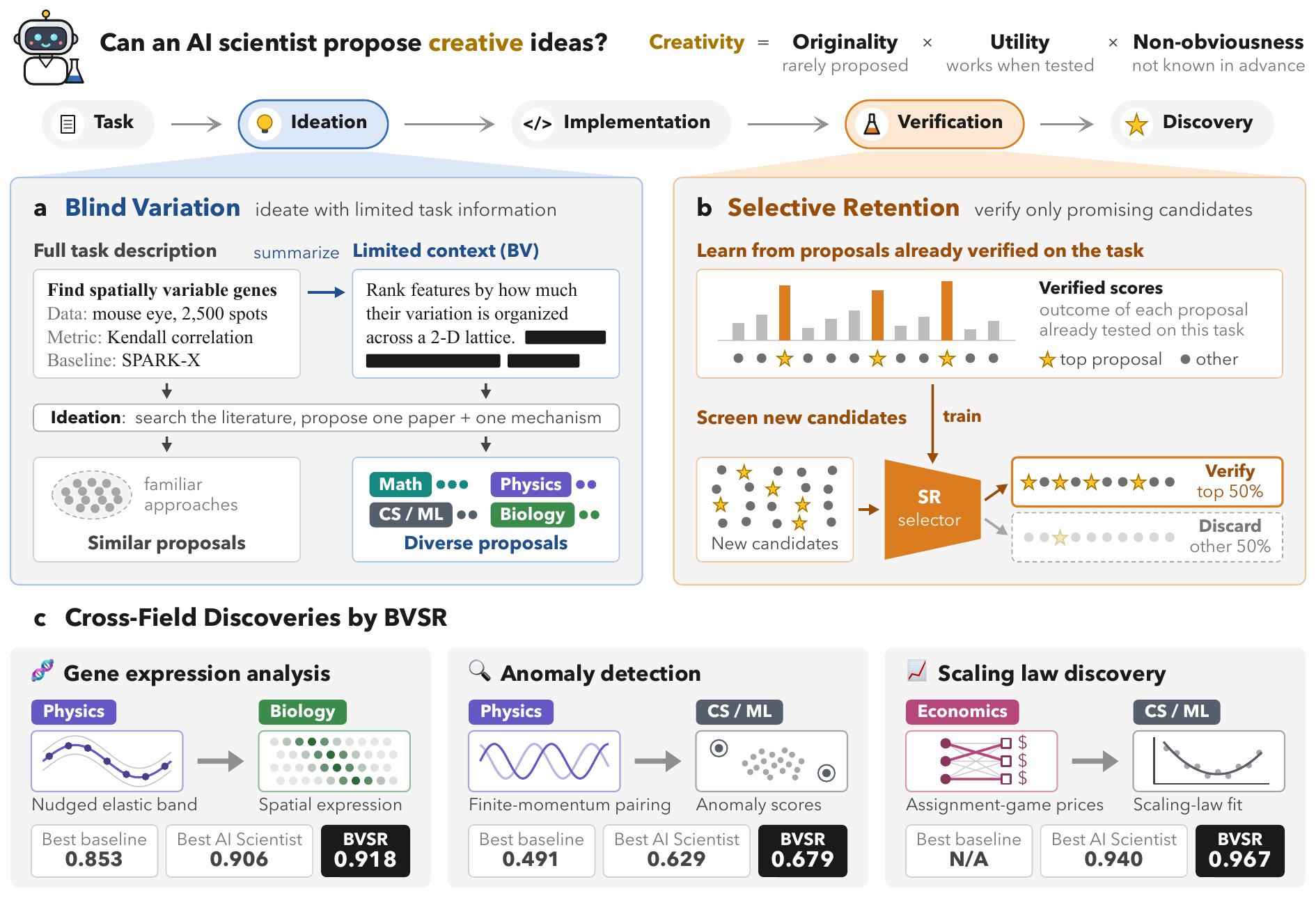}
    \vspace{-6pt}
    \caption{\textbf{Overview of \textcolor{figBlue}{Blind Variation} and \textcolor{figOrange}{Selective Retention} (BVSR).} (a) Limiting task information during idea generation encourages creative proposals. (b) Selective retention reduces the number of candidates requiring empirical verification while preserving top proposals. (c) BVSR's best programs borrow mechanisms from diverse fields, outperforming the best AI scientist on biology and ML tasks.}
    \label{fig:pipeline}
    \vspace{-10pt}
\end{figure}
\endgroup

\clearpage

\section{Introduction}
\label{sec:intro}

As large language models (LLMs) have made substantial gains in reasoning and long-horizon task execution, a growing body of work explores deploying AI agents for scientific discovery~\citep{aisci, cosci}. Existing studies demonstrate that AI agents can outperform the best human-designed solutions across a wide range of tasks, including mathematical proof~\citep{alphaproof}, mathematical discovery~\citep{funsearch}, optimization~\citep{alphaevolve}, and a subset of scientific problems~\citep{tttdiscover}.
However, strong performance alone does not establish the creativity of the proposed methods. Prior work has identified limitations in the novelty of AI-generated proposals, including a tendency to explore incremental modifications or extensions of existing approaches~\citep{aiscientistv2,codescientist}. A key challenge is therefore to develop AI agents that can reliably generate proposals that are creative. Addressing this challenge requires both assessing creativity rigorously and developing methods to improve it.

Since the 1950s, psychologists have systematically studied how creativity should be judged in humans~\citep{crea_1953, def_crea}. Their work identifies three defining properties of creative ideas: novelty, effectiveness, and non-obviousness~\citep{creativity}. These properties distinguish creativity from performance alone: an idea must work, but its originality and whether its success could have been anticipated also matter. This perspective is captured by a quantitative definition~\citep{quant_creativity}:\looseness=-2

\begin{equation}
\label{eq:creativity}
\text{Creativity} \;=\; \text{Originality} \times \text{Utility} \times \text{Non-obviousness} \;=\; (1-p)\cdot u\cdot(1-v),
\end{equation}

where $p$ is the probability of generating an idea, $u$ is its tested utility, and $v$ represents prior knowledge of its effectiveness. Originality captures whether an idea is uncommon rather than something that many others would propose. Utility reflects whether the idea actually works in practice. Non-obviousness concerns whether its effectiveness is known in advance. In this paper, we adopt this definition to assess the creativity of AI-generated proposals.

To improve the creativity of AI agents, prior work has relied mainly on prompt engineering, retrieval augmentation, and a variety of search methods: instructing the agent to be novel~\citep{analogical}, supplying it with literature as inspiration~\citep{literature_for_sci}, or iteratively refining and selecting among many candidates according to search heuristics~\citep{shinkaevolve}. Each of these approaches, however, has serious limitations. Prompt engineering yields diminishing returns on frontier models~\citep{prompt}. Literature is a valuable reference, but the literature an agent retrieves for itself suffers from the same lack of diversity as its own proposals~\citep{swiss, nova}. Heuristic search methods rely on a fine-grained signal, whereas most scientific problems offer at best a coarse metric of how well a finished proposal works~\citep{bes}. 

Motivated by these limitations, we draw on the theory of \textbf{Blind Variation (BV)} and \textbf{Selective Retention (SR)} to guide how AI agents generate and select proposals~\citep{bvsr}. This theory describes creative discovery as a process involving two complementary stages. The first is blind variation: scientists generate diverse ideas without foreknowledge of which will succeed, sometimes drawing inspiration from experiences unrelated to the immediate problem~\citep{constrain_crea, blind_human}. The second is selective retention: based on prior experience, they discard ideas that are clearly unreasonable and retain plausible ones for further investigation. Similarly, we find that in AI agents, providing too much task information during proposal generation causes the model to anchor on familiar approaches. As illustrated in Figure \ref{fig:pipeline}, our method on blind variation first limits the task information available during generation. We then use selective retention to learn, as a scientist does by working on a problem, from proposals already verified on that task and their empirical outcomes to identify promising candidates for further evaluation.

We first show that limiting task information during proposal generation can improve creativity. We provide an intuition for its effect on originality under information-induced anchoring, and validate its benefits empirically across five models and 12 scientific tasks. Across these settings, BV outperforms state-of-the-art prompting-based baselines in originality and non-obviousness, while generating a larger share of highly creative and useful proposals. We then develop SR to learn from previously verified proposals and their empirical outcomes, using a pairwise objective that separates high-utility proposals from the rest. SR retains top proposals more effectively than prompting-based methods, in-context learning, and supervised fine-tuning. Finally, we integrate BVSR into SimpleTES~\citep{simpletes}, an existing common adopted AI scientist workflow, where it largely improves discovery performance over the original baselines. \looseness=-2

\section{Preliminaries}
\label{sec:prelim}
An AI scientist attempts to solve an automated scientific discovery task $T$ by producing $n$ candidate methods. Each candidate $c$ is produced in two stages. An \emph{ideation} stage reads a problem context $I$ (possibly only part of the full problem description $S$) and returns an idea together with a reference paper supporting it. An \emph{implementation} stage then reads $S$ and the idea and writes the code. Both stages are driven by instruction prompts. We may vary the prompts and the context $I$. The task provides an evaluator that runs a candidate's code and returns a score $s_T(c)$.

\textbf{Measuring creativity.} Following \Eqref{eq:creativity}, each candidate is scored as $\mathrm{creativity}(c)=(1-p)\,u\,(1-v)$, with the three terms computed as follows.

\emph{Originality $1-p$.} $p(c)$ is the probability that the ideation stage, run again under the same context $I$ and prompt, produces the idea of $c$:
\begin{equation}
\label{eq:p}
p(c)=\Pr_{c'}\big[c'\equiv c\big],
\end{equation}
where $c'$ is an independent candidate from the same generator and $c'\equiv c$ denotes that the two express the same idea. In practice, we use an LLM to partition the $n$ candidates into groups expressing the same idea and estimate $p(c)$ by $|g(c)|/n$, where $g(c)$ is the group containing $c$. Because $1-p$ tends to be closer to $1$ when $n$ is large, we place originality on a log scale and define it as $1-\log|g(c)|/\log n$. Here, originality refers specifically to generative rarity, following the creativity literature, rather than novelty relative to the scientific literature.

\emph{Utility $u$.} We run the candidate's code under the task evaluator and normalize its improvement over the task's baseline score $s_T^{\mathrm{base}}$ by the best improvement achieved by any candidate on $T$ in our experiments, $s_T^{\mathrm{best}}$:
\begin{equation}
\label{eq:u}
u(c)=\max\!\left(\frac{s_T(c)-s_T^{\mathrm{base}}}{s_T^{\mathrm{best}}-s_T^{\mathrm{base}}},\,0\right).
\end{equation}

\emph{Non-obviousness $1-v$.} $v(c)$ measures how much evidence existed at proposal time that the idea would work. An LLM judge reads $S$ and the idea, and outputs $v(c)$, where $v=1$ means prior experience strongly suggests the mechanism will succeed, and $v=0$ means its success cannot readily be inferred from existing knowledge.

We assess the reliability of LLM judgments using PhD-level human experts. Human validation of both semantic grouping used for originality and non-obviousness is provided in Section~\ref{sec:human_validation}.

\textbf{Information level.}
We vary the information provided in context $I$ and quantify how much it reveals about the problem description $S$ using $\mathcal{V}$-usable information~\citep{v_use, under_v_use}, a model-dependent analogue of mutual information:
\begin{equation}
\label{eq:info}
\mathrm{info}(I)=H_{\mathcal V}(S)-H_{\mathcal V}(S\mid I),
\end{equation}
where $H_{\mathcal V}(S)$ and $H_{\mathcal V}(S\mid I)$ are the unconditional and conditional $\mathcal V$-entropies of $S$, respectively. They measure the uncertainty about $S$ for predictors in $\mathcal V$ without and with context $I$. In practice, we estimate these quantities using the negative log-likelihood of the same problem description $S$ under a language model, first with an empty context and then with $I$, using base-2 logarithms to express both values in bits. Their difference measures how much easier $S$ becomes to predict when $I$ is provided.  See Appendix~\ref{app:info} for details.

\section{Blind Variation: Limited Information Induces More Creativity}
\label{sec:blind}
In this section, we show that providing a model with too much task information during proposal generation steers it toward common, less novel approaches. This substantially reduces the originality and non-obviousness of its proposals and, in turn, limits the utility of its top proposals. As illustrated in Figure~\ref{fig:pipeline}(a), blind variation limits task information during idea generation to encourage exploration beyond familiar approaches.

\subsection{Motivation}
Recall that during proposal generation, the model reads a problem context $I$ and returns an idea along with a reference paper. We study how proposal creativity varies with the amount of information contained in $I$. 
While additional task information can strengthen task relevance, it also aligns independent generations around the same task-specific signal and reduces exploration.
We characterize this effect in the following proposition.
\begin{proposition}[Information--Exploration Tradeoff]
\label{thm:BV}
Under Assumption~\ref{ass:shared_signal}, the expected originality of generated proposals is non-increasing with task information:
\begin{equation}
\frac{d}{d\,\mathrm{info}(I)}
\mathbb{E}[o(c)]
\leq 0.
\end{equation}
Under a fixed generation budget, if higher-utility ideas are less associated with the shared task signal, increasing task information decreases the expected utility of the best generated proposal.
\end{proposition}
The assumptions and derivation are provided in Appendix~\ref{app:bv_theory}.

\subsection{Experimental Setup}
To test this empirically, we use Claude Opus 5 to summarize the full problem statement $S$, producing contexts $I$ with varying amounts of information for the same task. Appendix~\ref{app:info-examples} shows example contexts. We then evaluate proposal creativity under each context.  Further details on the models, benchmarks, and settings
 are provided in
Appendix~\ref{app:setup}. For each task, each model, and each information level, we generate 112 proposals.

\textbf{Benchmarks.}
We evaluate 12 tasks across four scientific domains.
(1) \emph{Biology.} We use four single-cell analysis tasks~\citep{sca}, namely cell-cell communication, label projection, perturbation prediction, and spatially variable genes.
(2) \emph{Astrodynamics.} We use the Cassini~\citep{cassini} and Voyager 2~\citep{voyager} trajectory design problems.
(3) \emph{Machine Learning.} We use unsupervised anomaly detection and post-hoc probability calibration from MLS Bench~\citep{mlsbench}, along with scaling-law discovery in pretraining and  U-shaped / double-descent regime from SLDBench~\citep{scalinglaw}.
(4) \emph{Neuroscience.} We use neural activity forecasting from ZAPBench~\citep{zapbench} with prediction horizons of 16 and 32.

\textbf{Models.}
We evaluate five mainstream LLMs to assess whether our findings generalize across model families. These include DeepSeek-V4-Flash, DeepSeek-V4-Pro~\citep{dsv4}, GLM-5.3 Flash, GLM-5.2~\citep{glm5}, and Kimi-K3~\citep{k3}. All models have reasoning enabled and access to OpenAlex~\citep{openalex} for literature search.

\subsection{Findings}
\begin{figure}[h]
    \centering
    \includegraphics[width=\linewidth]{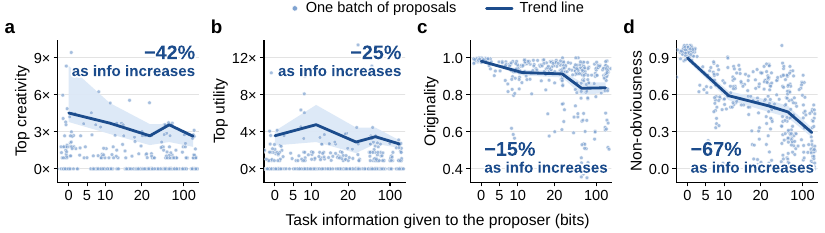}
    \caption{\textbf{Effects of task information on proposal creativity.} Each dot summarizes one batch of proposals for a benchmark, information level, and model. The trend line follows the 95th percentile (a, b) or the mean (c, d) of the dots. Percentages give its change from 0 to 140 bits. As information increases, the relative frequency of proposals ranked in the benchmark's top 1\% for creativity or utility generally decreases, while mean originality and mean non-obviousness decline substantially.}
    \label{fig:main}
\end{figure}

Figure~\ref{fig:main} shows how creativity, utility, originality, and non-obviousness change as the amount of task information increases. Each dot represents the proposals generated by one model for one benchmark at one information level.

Since AI scientist tasks place particular value on highly creative and highly useful methods, we first examine how top creativity and top utility vary with information level. We focus on proposals ranked in the top 1\% within each benchmark, according to creativity and utility, respectively. The vertical position of each dot indicates a relative frequency, computed as the fraction of its proposals that fall into this top 1\%, divided by 1\%. We find that models generally produce a larger share of these top-ranked proposals at lower information levels.

We next examine whether originality and non-obviousness follow a similar pattern. We plot mean originality and mean non-obviousness against information level and find that both decline substantially as the amount of task information increases.
Across proposals generated by five models on 12 tasks, these results provide systematic evidence that increasing task information reduces the relative frequency of highly creative proposals, supporting the conclusion that limited information can induce greater creativity. We call this approach of improving model creativity by limiting the information provided \emph{Blind Variation} (BV).

\subsection{Baseline Comparison}
Prior work improves AI-scientist creativity by modifying prompt instructions. We compare against three such baselines: analogical reasoning, cross-domain search, and no-domain search~\citep{analogical}. We further add one control: BV with the full task description, which separates the effect of limited information from that of our prompt. All methods share the same proposer models, candidate budget, and reasoning configuration. For each model and task, we generate 112 proposals and report in Table~\ref{tab:creativity_gap_ours} the top creativity and top utility, defined here as the mean creativity and utility of the top 1\% of proposals, together with mean originality and mean non-obviousness.

\begin{table}[h]
    \centering
    \caption{\textbf{Comparison of BV with baselines.} The highest is shown in bold. Values are mean $\pm$ SE.}
    \label{tab:creativity_gap_ours}
    \setlength{\tabcolsep}{5pt}
    \renewcommand{\arraystretch}{1.15}
    \resizebox{\linewidth}{!}{
    \begin{tabular}{lcccc}
        \toprule
        Method &
        \shortstack{Creativity ($\uparrow$)} &
        \shortstack{Utility ($\uparrow$)} &
        \shortstack{Originality ($\uparrow$)} &
        \shortstack{Non-obviousness ($\uparrow$)} \\
        \midrule
        BV (Limited Information)
        & $\mathbf{0.530 \pm 0.043}$
        & $\mathbf{0.571 \pm 0.041}$
        & $\mathbf{0.955 \pm 0.002}$
        & $\mathbf{0.713 \pm 0.012}$ \\
        BV (Full Information)
        & $0.462 \pm 0.045$
        & $0.524 \pm 0.043$
        & $0.926 \pm 0.004$
        & $0.609 \pm 0.015$ \\
        Analogical Reasoning
        & $0.429 \pm 0.049$
        & $0.508 \pm 0.050$
        & $0.914 \pm 0.007$
        & $0.492 \pm 0.037$ \\
        Cross-Domain
        & $0.369 \pm 0.052$
        & $0.531 \pm 0.045$
        & $0.871 \pm 0.012$
        & $0.263 \pm 0.020$ \\
        No-Domain
        & $0.117 \pm 0.025$
        & $0.530 \pm 0.050$
        & $0.598 \pm 0.020$
        & $0.104 \pm 0.015$ \\
        \bottomrule
    \end{tabular}
    }
\end{table}

BV achieves the highest creativity, utility, and non-obviousness. On the same tasks, the full task description costs the same prompt a large part of its advantage over the baselines: creativity falls from 0.530 to 0.462 and non-obviousness from 0.713 to 0.609, indicating that limiting task information is a major source of BV's gains. 
 \section{Selective Retention: Learning from Scientific Experience}
\label{sec:rent}
Blind variation deliberately reduces task-specific guidance during idea generation, but scientific discovery should not ignore  what is learned by working on the task.  Just as scientists develop judgment about a problem by testing ideas on it, we formulate selective retention as learning from experience on the task itself. Each proposal previously verified on the task provides a pair consisting of a candidate and its empirical outcome. From these proposal–outcome pairs, we learn a scoring function that decides which new candidates for that task should be retained for expensive execution. As illustrated in Figure~\ref{fig:pipeline}(b), selective retention (SR) uses this learned judgment to reduce the number of candidates requiring empirical verification while preserving promising proposals.

\subsection{Motivation}
Verification can be time-consuming in many scientific tasks: wet-lab experiments may take months, and machine learning experiments may require substantial GPU computation~\citep{mlexp}. A natural approach is to prompt an LLM to inspect candidates. However, as our comparison below shows, these prompting-based approaches offer limited preservation of top candidates. This motivates learning a screener rather than relying exclusively on prompted judgments.
The learning objective must also reflect the purpose of screening. Pointwise supervised fine-tuning can train a model to predict each candidate's performance, but its regression loss does not explicitly prioritize the preservation of top candidates. Selective retention has a more targeted requirement: high-value candidates should remain above the discard region. Precise prediction and ordering are secondary. We therefore focus on separating the top proposals from the remaining ones.
We first argue that identifying clearly ineffective methods is much easier than identifying the most effective ones. 

\begin{proposition}[\textbf{Selective Retention Advantage}]
\label{prop:SR}
Under the mild Utility-Gap Assumption \ref{ass:utility_gap}, if
$\Delta_{\mathrm{SR}}(d) > \Delta_{\mathrm{best}}$,
then a comparison across the selective-retention boundary is more accurate than distinguishing the best candidate from its closest competitor. Formally, we have
\[
P_{\mathrm{correct}}\!\left(\Delta_{\mathrm{SR}}(d)\right)
>
P_{\mathrm{correct}}\!\left(\Delta_{\mathrm{best}}\right).
\]
\end{proposition}

We then argue SR reduces the verification time required to achieve the same performance, provided that its probability of discarding top proposals is significantly lower than that of random discarding.

\begin{proposition}[\textbf{Safe Discard Threshold}]
\label{thm:safe_discard}
Let $d\in[0,1]$ denote the overall discard ratio and $\beta_q(d)$ the expected fraction of top-$q$ proposals discarded by SR. For a tolerated loss $\epsilon$, define
\[
d_q^\star(\epsilon)=\sup\{d:\beta_q(d)\leq\epsilon\}.
\]
Then $d_q^\star(\epsilon)$ is the largest discard ratio that limits the expected loss of top-$q$ proposals to at most $\epsilon$, and therefore minimizes the number of proposals requiring verification under this constraint.
\end{proposition}

\begin{remark}
For random discarding, $\beta_q^{\mathrm{rand}}(d)=d$, so $d_{\mathrm{rand}}^\star(\epsilon)=\epsilon$. Hence, whenever $d_q^\star(\epsilon)>\epsilon$, selective retention permits a larger discard ratio than random under the same tolerated loss of top proposals.
\end{remark}

The assumptions and derivations for both propositions are provided in Appendices~\ref{app:sr_utility_gap} and~\ref{app:sr_safe_discard}.

\subsection{Learning to Select Top Proposals}
We implement selective retention using a language model with a scalar scoring head. Given a task specification $S$ and a candidate $x$, the model produces a score $s=f_\theta(S,x)$.
For a training run, let $\mathcal{T}$ contain its top-5\% utility candidates. We include all candidates in $\mathcal{T}$ and uniformly sample candidates $\mathcal{B}$ from the remainder. We apply a pairwise logistic loss~\citep{ranknet}, restricted to comparisons between these two groups:
\begin{equation}
\mathcal{L}_{\mathrm{SR}}
= \sum_{i\in\mathcal{T}}\sum_{j\in\mathcal{B}}
\log\!\left(1+\exp\!\left[-(s_i-s_j)\right]\right).
\label{eq:sr_top_rest}
\end{equation}
Each comparison can be interpreted as a unit of scientific experience: given two previous proposals, the model learns which one was more worthy of empirical investment. Aggregating many such comparisons 
trains the selector to exploit past outcomes when screening its future candidates.

\textbf{Experimental Setup.}
To evaluate the effectiveness of our method, we use 5,783 proposals generated by the five proposer models on the 12 tasks in Section~\ref{sec:blind}. Our training dataset consists of a separate set of 40,733 proposals on the same tasks. Implementation details 
are provided in Appendix~\ref{app:setup-sr}. To assess whether the model can identify low-quality proposals, we ask it to rank the candidates and discard those ranked lowest. We vary the discard ratio and measure the fraction of top candidates kept. Higher curves indicate better preservation.
All screeners receive the same task specification and candidate implementation. We compare SR with four zero-shot prompting-based judges: \textit{\textbf{Pointwise}} judge assigns absolute scores, \textit{\textbf{Setwise}} judge repeatedly identifies weak candidates in groups, \textit{\textbf{Listwise}} judge aggregates sliding-window rankings in the style of RankGPT~\citep{rankgpt}, and \textit{\textbf{Tournament}} judge uses six rounds of Swiss-system pairwise comparisons~\citep{swiss}. The in-context learning (\textit{\textbf{ICL}}) judge also sees four similar verified proposals and their outcomes from the SR training set. We also compare with \textbf{\textit{SFT}} using the same backbone, trained with mean-squared error.

\begin{figure}[!t]
    \centering
    \includegraphics[width=\linewidth]{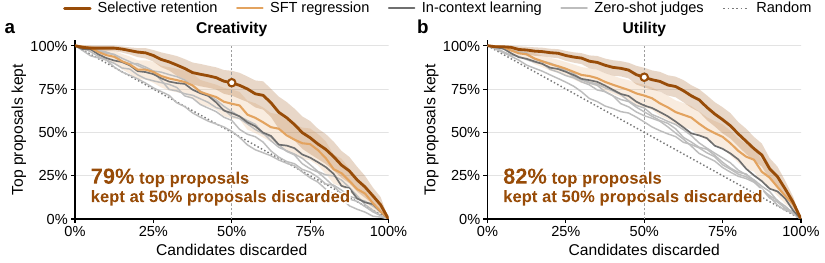}
    \caption{\textbf{Preserving top candidates during selective retention.} Fraction of top proposals kept as discard ratio varies. Selective retention keeps the most top candidates, ahead of SFT regression and prompt-based judges.}
    \label{fig:swiss}
\end{figure}

\textbf{Findings.}
Figure~\ref{fig:swiss} shows that learned SR preserves top candidates much more effectively than prompting-based screeners. The in-context learning baseline, despite seeing verified outcomes of similar proposals, performs no significantly better than zero-shot judges.
SFT improves preservation over many prompt-based methods. Nevertheless, it remains less effective than SR. This result is consistent with the motivation that accurate prediction throughout the distribution is a different objective from keeping exceptional candidates out of the discard set.
At a discard ratio of 50\%, SR preserves 79\% of the top-5\% proposals by creativity and 82\% of the top-5\% proposals by utility. We therefore use this SR model to discard the lowest-ranked 50\% of proposals in later experiments.

\section{Applications}

\subsection{Cross-field Discovery}

\begin{figure}[h]
    \centering
    \includegraphics[width=\linewidth]{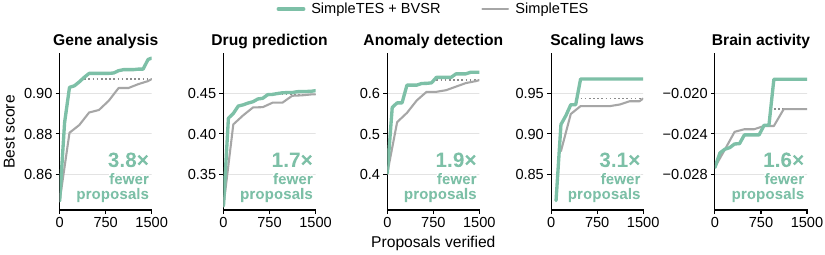}
    \caption{\textbf{Integrating BVSR into SimpleTES.} Best task score as a function of verification steps on five scientific tasks.  BVSR improves final score and reaches strong solutions with fewer steps. Numbers give how many times fewer verified proposals SimpleTES + BVSR needs to reach SimpleTES's final score (dotted line).}
    \label{fig:simpletes}
\end{figure}

Beyond the gains of BV and SR individually, we evaluate whether combining information-limited generation with experience-based selection improves end-to-end scientific discovery. We evaluate its effectiveness by integrating it into SimpleTES~\citep{simpletes}.
We compare SimpleTES with and without BVSR on five tasks introduced earlier: gene expression analysis (spatially variable genes), drug response prediction (perturbation prediction), anomaly detection, scaling law discovery, and brain activity prediction (ZAPBench horizon 16).
We use Kimi-K3 for both configurations. Following the SimpleTES setup, we run four parallel search trajectories, each generating 40 proposals per iteration, and run for 10 iterations.
Figure~\ref{fig:simpletes} plots the best score against the cumulative number of empirically verified proposals.
BVSR achieves higher final scores on all five tasks and exceeds the original workflow's final score with fewer verification steps. These results show that BVSR can improve discovery performance when incorporated into existing AI scientist workflows.

\subsection{Case Study}
\label{app:case}

We describe three example best programs found by BVSR in Table~\ref{tab:case}. Each surpasses the best program found by SimpleTES under the same budget and the strongest existing baselines on their tasks.

\begin{table}[h]
\centering
\small
\caption{Best programs of our methods and of SimpleTES on example tasks (higher is better). Best baseline: SPARK-X for gene expression analysis, Isolation Forest for anomaly detection.}
\label{tab:case}
\resizebox{0.65\textwidth}{!}{%
\begin{tabular}{@{}lccc@{}}
\toprule
Task  & Ours & SimpleTES & Best Baseline \\
\midrule
Gene Expression Analysis   &  \textbf{0.918} & 0.906 & 0.853~\citep{sparkx} \\
Anomaly Detection  &  \textbf{0.679} & 0.629 & 0.491~\citep{isolation} \\
Scaling Law Discovery   &  \textbf{0.967} & 0.940 & N/A \\
\bottomrule
\end{tabular}%
}
\end{table}

\textbf{Spatially variable genes via the nudged elastic band.}
The task is to rank genes by how strongly their expression is organized in space on a tissue section.
The program borrows the nudged elastic band method from transition-state theory \citep{jonsson1998neb}, which finds a minimum-energy path by relaxing a chain of images under two projected forces: the true force perpendicular to the path and a spring force along it.
It takes the heat-flow trajectory of each gene's expression field on the tissue graph, from the observed field to the uniform one, and computes at every image the component of a robust energy gradient perpendicular to the path.
For pure noise, the heat-flow path is already the minimum-energy path, and this component vanishes; spatial patterns bend the path, so the perpendicular force grows with pattern strength and becomes the gene's score.
It reaches 0.918, against 0.906 for SimpleTES and 0.853 for SPARK-X, the best method on OpenProblems.

\textbf{Anomaly detection via finite-momentum pairing.}
The task asks for an anomaly detector with at most 10,000 parameters, scored on four tabular datasets.
The program transplants the nonuniform (FFLO) superconducting state \citep{larkin1964fflo}, in which two species with mismatched Fermi surfaces still pair by giving each pair a nonzero momentum, so that the order parameter oscillates in space.
Its two ``species'' are two scores that often disagree: marginal tail evidence from empirical distribution functions, and geometric evidence from a robust covariance and a Gaussian mixture of the normal mass.
Instead of averaging them, it pairs them with an oscillating modulation along the direction in which they disagree most, relaxes the recovered pairing signal on a nearest-neighbour graph and adds it to the combined score.
With no trained network it reaches 0.679, against 0.629 for SimpleTES and 0.491 for Isolation Forest.

\textbf{Scaling law via assignment-game prices.}
The task is to find a formula with at most 26 parameters for the final training loss as a function of learning rate, batch size, data size and model size.
The program draws on the assignment game of \citet{shapley1971assignment}, whose optimal matching is a linear program with competitive prices as dual variables.
It solves one fixed, entropy-regularized assignment problem between two grids in the (log learning rate, log batch size) plane and uses the resulting price field as a fixed surface that reshapes the loss basin near the optimal learning rate and batch size. The law adds power laws in data and model size and a convex bowl around a scale-dependent optimum, and fits only two parameters.
It reaches a 0.967 held-out $R^2$.

\subsection{Ablation Study}

\sbox{\ablationfigurebox}{%
    \begin{minipage}[t]{0.40\linewidth}
        \vspace{0pt}
        \centering
        \includegraphics[width=\linewidth]{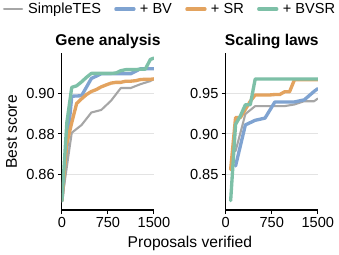}
        \captionof{figure}{\textbf{Ablation study.}
        Best task score versus verification steps for four
        SimpleTES configurations. Higher is better.}
        \label{fig:abl}
    \end{minipage}%
}

\noindent
\begin{minipage}[t][\dimexpr\ht\ablationfigurebox+\dp\ablationfigurebox\relax][s]{0.56\linewidth}
    \vspace{0pt}
    To assess the individual contributions of BV and SR,
    we compare four configurations: the original SimpleTES, SimpleTES with BV alone, SimpleTES with SR alone,
    and the full BVSR. We evaluate all four configurations
    on gene expression analysis and scaling law discovery, and track the best task score against verification steps.
    Figure~\ref{fig:abl} shows that BV and SR each improve the final score over SimpleTES, and that combining them works best on both tasks. On gene expression analysis, BV mainly raises the final score and SR mainly speeds up discovery, while BVSR achieves a larger gain than either. On scaling law discovery, BVSR again reaches a higher score earlier than BV or SR. Because BVSR improves upon the stronger component on each task, the two components are complementary.
\end{minipage}\hfill
\usebox{\ablationfigurebox}
\par

\subsection{Human Validation}
\label{sec:human_validation}

We conduct two human studies to validate the LLM judgments underlying originality and non-obviousness. The former relies on grouping semantically equivalent proposals to estimate their frequencies, while the latter assesses whether prior knowledge suggests that a proposal will work. Detailed protocols are provided in Appendix~\ref{app:human_validation}.

\textbf{Originality.}
Five PhD-level researchers each annotate 30 proposal pairs, yielding 150 pairs in total. The sample is balanced between pairs grouped together and pairs assigned to different groups by the LLM, and covers multiple tasks, proposer models, and generation methods. Evaluators judge whether the proposals express the same underlying mechanism.  84\% of pairs grouped together by the LLM are also judged equivalent by humans (95\% CI $[0.74,0.92]$), while 73\% of pairs assigned to different groups are judged distinct (95\% CI $[0.61,0.82]$). These results support the semantic-equivalence judgments used to estimate originality.

\textbf{Non-obviousness.}
Five PhD-level researchers evaluate 150 high-quality proposals from BV and Analogical Reasoning (AR) under blinded conditions. All sampled proposals achieve verified scores above the best baseline for their respective tasks. Evaluators see the task name and summaries of the source paper and proposed method. The summaries are rewritten in a uniform style, and method identifiers are removed. Evaluators assess how readily a proposal's effectiveness could be inferred from the paper and their own knowledge of the literature, using a 1--5 scale with higher scores indicating greater non-obviousness.
Human ratings correlate with the automatic non-obviousness score $1-\hat{v}$ at $r=0.56$ (95\% CI $[0.45,0.67]$). BV proposals receive a mean rating of $3.64$, compared with $2.97$ for AR, a difference of $0.67$ points (95\% CI $[0.33,0.99]$, $p=0.0006$). These results support the automatic non-obviousness score and show that BV's advantage persists under expert judgment.

\subsection{Domains of Inspiration}
\begin{figure}[h]
    \centering
    \includegraphics[width=0.95\linewidth]{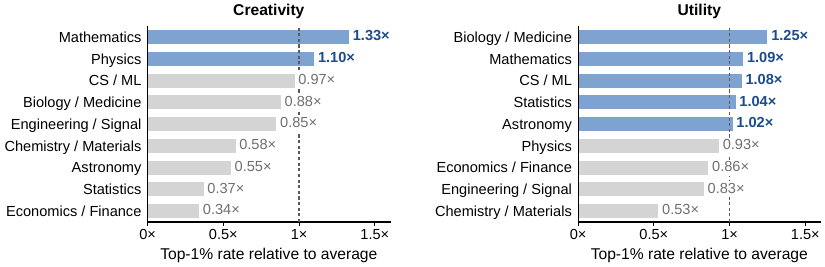}
    \caption{\textbf{Proposal quality across domains.}
    Normalized top-1\% fractions by creativity and utility.}
    \label{fig:domain}
\end{figure}

During proposal generation, models search for papers as
sources of inspiration. To examine which scientific domains
contribute more to top proposals, we sample 20,000
proposals and group them by the domains of their reference
papers. For creativity and utility separately, we compute
the fraction of proposals from each domain that enter
their respective task's top 1\%, divided by the corresponding
fraction across all proposals. A value above 1
indicates an above-average likelihood of producing a top
proposal. Figure~\ref{fig:domain} shows that mathematics
has the highest normalized fraction for top creativity
(1.33), while biology / medicine and mathematics lead in top utility
(1.25 and 1.09).

\section{Related Work}
\label{sec:related_work}

\textbf{AI Scientists.}
Recent work increasingly connects language models to scientific workflows.
Coscientist integrates research planning with laboratory automation, while Agent Laboratory and Virtual Lab organize research through specialized agents and human guidance~\citep{boiko2023,schmidgall2025agentlaboratory,swanson2025virtuallab}.
The AI Scientist extends automation across ideation, experimentation, and manuscript preparation~\citep{aisci}.
 AlphaEvolve and SimpleTES organize proposal generation and execution~\citep{alphaevolve,simpletes}.
These systems provide capable discovery workflows, but their effectiveness also depends on the candidates they explore.
Our work studies how task information shapes proposal generation and how empirical outcomes from previous proposals can guide which new candidates are selected for verification.

\textbf{Creativity in Humans.}
Associative theories explain creativity through connections between previously remote concepts~\citep{mednick1962associative,finke1992creative}.
Studies show that familiar categories and example solutions can constrain subsequent ideas~\citep{ward1994structured,jansson1991design}.
Research on incubation further suggests that temporarily setting a problem aside can improve creative problem solving~\citep{sio2009incubation}.
Idea selection presents a separate challenge: people often favor feasible and desirable candidates at the expense of originality~\citep{rietzschel2010selection}.
Together with Campbell's blind-variation-and-selective-retention theory~\citep{bvsr}, these findings help explain the relationship between information and creativity.
We investigate whether reducing task information during ideation can encourage exploration in AI scientists.

\textbf{Creativity in AI.}
Computational creativity distinguishes combinational, exploratory, and transformational processes~\citep{boden1998creativity}.
Recent work evaluates LLM creativity through divergent-thinking tasks and the diversity of repeated generations~\citep{koivisto2023best,noveltybench}.
Beyond text generation, Creative Agents incorporates imagination into embodied planning for open-ended construction tasks~\citep{creativeagents}.
Methods for diversifying outputs include verbalized sampling~\citep{verbalized}, which elicits distributions over responses, and analogical reasoning~\citep{analogical}, which transfers relational structures across scientific domains.
Our approach adds task information level as a distinct control for creative generation.
\section{Conclusion}
\label{sec:conclusion}

We studied how information should be used at different stages of automated scientific discovery. Our results suggest that more information is not uniformly beneficial. During idea generation, detailed task information can anchor language models toward familiar mechanisms and reduce the probability of generating creative proposals. Blind Variation mitigates this effect by limiting task information.
After generation, empirical experience earned on the task can guide which new candidates deserve expensive verification. Selective Retention learns this decision from proposal–outcome data.
Together, these findings suggest a broader design principle for AI scientists: separate the information used for exploration from the experience used for selection.

\bibliography{iclr2027_conference}
\bibliographystyle{plainnat}

\clearpage
\appendix

\etocdepthtag.toc{appendix}
\etocsettocstyle{\section*{Appendix - Table of Contents}}{}
\tableofcontents

\clearpage
\section{Details about Information Level}
\label{app:info}

By definition, the two $\mathcal V$-entropies in \eqref{eq:info} are the shortest code lengths that any predictor in $\mathcal V$ achieves for the problem description $S$, with and without the context $I$~\citep{v_use}:
\begin{equation}
\label{eq:info-def}
H_{\mathcal V}(S) = \inf_{f\in\mathcal V} -\log_2 f[\varnothing](S), \qquad
H_{\mathcal V}(S\mid I) = \inf_{f\in\mathcal V} -\log_2 f[I](S).
\end{equation}
Here $\mathcal V$ should be the family of models that actually read the context, i.e.\ the large language models used as proposers, so the direct estimate would score $S$ with each proposer and take the minimum code length.
In practice this is not feasible since the proposers are large mixture-of-experts models, and the log-likelihood such a model assigns to a fixed text is not stable, because it varies with the compute environment.

We therefore use three off-the-shelf small dense models, Qwen3-8B~\citep{qwen3}, Gemma-3-4B-it~\citep{gemma3} and Llama-3.1-8B-Instruct~\citep{llama3}, and average their estimates.
$S$ is the benchmark's own description of the task.
Each model $M$ reads $I$ followed by $S$, and we sum the base-2 log-probabilities of the tokens of $S$, which gives $\log_2 P_M(S\mid I)$; $\log_2 P_M(S)$ is obtained from the same text with $I$ left empty.
The estimate is
\begin{equation}
\label{eq:info-est}
\widehat{\mathrm{info}}(I) \;=\; \frac{1}{3}\sum_{M} \Big[\log_2 P_M(S\mid I) - \log_2 P_M(S)\Big],
\end{equation}

Under a fixed evaluation setup, these small dense models yield reproducible token log-probabilities, enabling stable estimation of information levels. Beyond stability, we also examine whether the estimates depend on the choice of dense models. 

To this end, we scored 32 example contexts using 10 open-weight models: our three selected models, along with Llama-3.2-3B-Instruct~\citep{llama3}, Mistral-7B-v0.3, Qwen2.5-7B, Qwen3-4B~\citep{qwen3}, Qwen3.5-4B, Qwen3.5-9B, and Gemma-3-12B-it~\citep{gemma3}. We computed the Pearson correlation between each pair of models' estimates across the contexts.

\begin{figure}[h]
    \centering
    \includegraphics[width=0.6\linewidth]{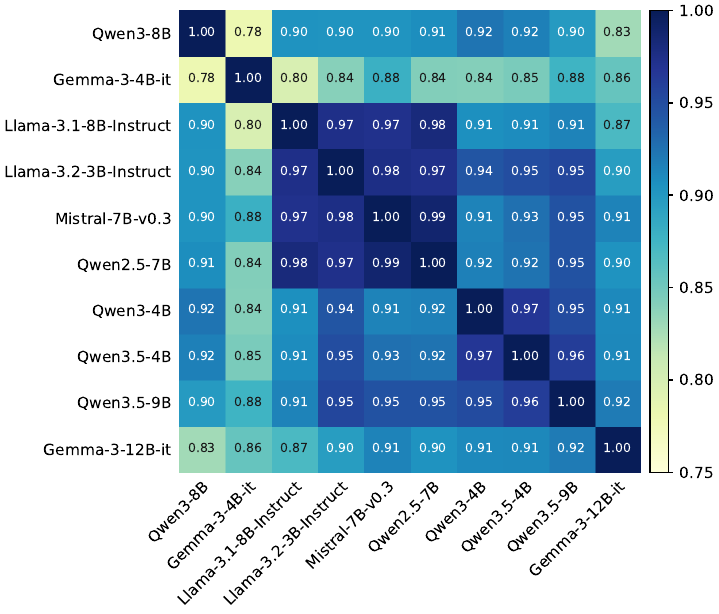}
    \caption{\textbf{Agreement across models on estimated information levels.} Each cell shows the Pearson correlation between two models' estimates across 32 contexts. Bold model names indicate the three models used in our estimator.}
    \label{fig:info-correlation}
\end{figure}

Figure~\ref{fig:info-correlation} shows strong agreement across models, with a median Pearson correlation of $r=0.91$ across the 45 model pairs. Furthermore, the mean estimate from our three selected models correlates with the mean across all 10 models at $r=0.99$. These results indicate that our estimates are robust to the choice of dense models.

\section{Experimental Setup}
\label{app:setup}

This section describes the models, the twelve discovery tasks, and the settings of blind variation and selective retention together with every baseline they are compared with.
Prompts are collected in Appendix~\ref{app:prompts}.

\begin{table}[h]
\centering
\footnotesize
\caption{The twelve discovery tasks across four scientific domains.
The time limit is the CPU wall-clock evaluation limit per candidate
(for \textsc{ZAPBench}, the GPU training budget on an RTX PRO 6000).}
\label{tab:setup-tasks}

\begin{tblr}{
    width = \linewidth,
    colspec = {
        Q[c,m]
        Q[l,wd=0.155\linewidth]
        X[2.8,l]
        X[1.8,l]
        X[1.4,l]
        r
    },
    colsep = 3pt,
    rowsep = 3pt,
    column{1} = {
        halign=c,
        valign=m,
        leftsep=3pt,
        rightsep=3pt
    },
    column{2-6} = {valign=t},
    column{6} = {rightsep=0pt},
    vline{2} = {0.4pt},
    cell{2}{1}  = {r=4}{halign=c,valign=m},
    cell{6}{1}  = {r=4}{halign=c,valign=m},
    cell{10}{1} = {r=2}{halign=c,valign=m},
    cell{12}{1} = {r=2}{halign=c,valign=m},
}
\toprule
Domain & Task & Data and Setup & Task Score & Reference Baseline & Time Limit \\
\midrule
\rotatebox[origin=c]{90}{Biology}
& \textsc{SVG}
& DBiT-seq mouse eye; 2,500 spots, 1,050 genes
  (50 genes $\times$ 21 variants)
& Kendall correlation in variant groups, scaled
& SPARK-X 0.8531
& 900\,s \\
& \textsc{LabelProj}
& GTEx v9; 197,618 reference and 11,508 query nuclei, 53 cell types
& Accuracy and three F1 scores, scaled
& scANVI+\allowbreak scArches 0.773
& 2,400\,s \\
& \textsc{PertPred}
& NeurIPS 2023 PBMC; 402 training, 151 query profiles; 5,317 genes
& Five row-wise metrics, scaled
& NN, pseudo-labels 0.465
& 1,500\,s \\
& \textsc{CCC}
& TNBC; 42,512 cells, 29 cell types; 812 truth pairs, 166 positive
& AUPRC and top-5\% odds ratio, scaled
& Product specificity 0.120
& 2,400\,s \\
\midrule
\rotatebox[origin=c]{90}{Machine Learning}
& \textsc{MLS-Anomaly}
& Cardio, Thyroid, Satellite, Shuttle; $\le 10{,}000$ parameters
& AUROC, F1 vs.\ baselines; geometric mean
& Isolation Forest 0.4912
& 300\,s \\
& \textsc{MLS-Calibration}
& Four fixed classifiers on MNIST, Fashion-MNIST, Madelon, Breast Cancer
& ECE, Brier, NLL vs.\ baselines; geometric mean
& Temperature scaling 0.2935
& 180\,s \\
& \textsc{SL-LR/BS}
& Loss vs.\ learning rate, batch, data, model size;
  2,702/117 points; $\le 26$ parameters
& Held-out $R^2$
& 0
& 1,500\,s \\
& \textsc{SL-U}
& U-shaped Brier score vs.\ log compute; 9 groups;
  389/127 points; $\le 6$ parameters per group
& Held-out $R^2$
& 0
& 1,500\,s \\
\midrule
\rotatebox[origin=c]{90}{Astrodynamics}
& \textsc{Cassini}
& Earth to Saturn; flybys allowed at Venus, Earth, Mars, Jupiter
& $2t/(t+\Delta v)$, $t = 1$\,km/s
& Initial program 0.2786
& 300\,s \\
& \textsc{Voyager-2}
& Earth to Neptune; flybys allowed at Jupiter, Saturn, Uranus
& $2t/(t+\Delta v)$, $t = 2$\,km/s
& Initial program 0.6610
& 300\,s \\
\midrule
\rotatebox[origin=c]{90}{Neuroscience}
& \textsc{ZAPBench-16}
& 71,721 neurons, 9 conditions; 4 context, 16 forecast frames
& $-$MAE
& Mean of context $-0.0289$
& 600\,s \\
& \textsc{ZAPBench-32}
& As above, 32 forecast frames
& $-$MAE
& Mean of context $-0.0335$
& 600\,s \\
\bottomrule
\end{tblr}
\end{table}

\subsection{Models}
\label{app:setup-models}

Proposals are generated by five LLMs: Kimi-K3, GLM-5.2, GLM-5.3-Flash, DeepSeek-V4-Flash and DeepSeek-V4-Pro.
Originality and non-obviousness (Section~\ref{sec:prelim}) are estimated by two OpenAI models. For originality, GPT-5.6-Sol reads all mechanism sentences and groups those that express similar ideas.
For non-obviousness, GPT-5.6-Luna reads the task description, the chosen paper, and the proposal, and returns $\hat v(x)\in[0,1]$.
The SR selector is a fine-tuned Qwen3.5-9B-Base model.

\subsection{Datasets}
\label{app:setup-data}

We use twelve discovery tasks from five benchmark families (Table~\ref{tab:setup-tasks}): four single-cell tasks~\citep{sca}, two tasks from MLS-Bench~\citep{mlsbench}, two scaling-law tasks from SLD-Bench~\citep{scalinglaw}, two gravity-assist trajectory tasks, and whole-brain activity forecasting from \textsc{ZAPBench}~\citep{zapbench} at horizons 16 and 32.
Every task asks for one executable program with a fixed entry point, which the benchmark's own evaluator runs end to end to produce the score.

\subsection{Blind Variation Setup}
\label{app:setup-bv}

In blind variation, the model receives limited task information to generate proposals, which are subsequently implemented as code. To vary the information available during proposal generation, we use Claude Opus 5 to generate four contexts with different amounts of task information for each task. Appendix~\ref{app:prompts_bv} provides the prompts used in BV, along with the prompt used to generate these contexts with Claude Opus 5.

We compare blind variation with three baselines: analogical reasoning, cross-domain search, and no-domain search, using the official prompts provided by \citet{analogical}. In no-domain search, the model independently searches for papers it considers most helpful. In cross-domain search, the model is instructed to search for papers across domains, while in analogical reasoning, it is instructed to identify relevant papers through analogy.
 BV (Full Information) always uses the full task description $S$ as its context. 

\subsection{Selective Retention Setup}
\label{app:setup-sr}

Selective retention ranks the candidates of a pool and discards the lowest-ranked fraction.
The ranking model is Qwen3.5-9B-Base~\citep{qwen3}, fully fine-tuned with a zero-initialized linear head that maps the final hidden state of the last input token to a score $f_\theta$, of which only the order is meaningful.
Its input is the task description followed by the candidate proposal.

We compare the selector with six ranking baselines.
The SFT baseline fine-tunes the same model on the same input but regresses, by mean squared error, the standardized within-run percentile rank of the verified score.

We also include four LLM judge baselines. Here, each cell is ranked by its own proposer.
\emph{Pointwise} scoring asks for an integer from 0 to 100 per candidate, with ties broken at random.
The \emph{tournament} plays six Swiss-system rounds of pairwise comparisons and ranks candidates by points, then by Buchholz score (the sum of the opponents' points), then at random.
\emph{Setwise} ranking asks which three of ten implementations to discard, over ten random partitions of the cell, and ranks candidates by how rarely they are named.
\emph{Listwise} ranking uses the sliding window of \citet{rankgpt} and averages each candidate's position over three random initial orders.

The in-context learning (ICL) baseline gives the judge access to the same training dataset as SR.
For each test candidate, we give four similar proposals from the SR training set, which covers the same tasks and is disjoint from the test set.
Each demonstration shows the implementation and its within-run percentile rank of the verified score.
The judge then scores the test candidate from 0 to 100 as in Pointwise.

\begin{promptbox}[mblueL]{mblueD}{Pointwise judge}{box:sr-pointwise}
You are deciding how worth it is to spend an expensive evaluation on ONE candidate IMPLEMENTATION for a task. It is real code that will be executed on the task's data and scored by the task's metric. Judge how likely the implementation is to (1) run without error and produce a valid output of the required form, and (2) score well on the task's metric.

TASK SPECIFICATION (data/inputs provided, required output, metric):
<task specification>

IMPLEMENTATION:
```python
<candidate code>
```

Read the implementation carefully. Consider: does it correctly consume the provided data/inputs; does it actually compute something aligned with the task's objective (not a trivial constant, a degenerate output, or leakage of the target values); is it likely to run without crashing; and does its approach plausibly beat the baseline. Judge the ACTUAL code, not the comments' claims. Ignore length/verbosity.

First THINK briefly about the points above (a few sentences). THEN give your FINAL ANSWER as one JSON object in a ```json code fence:
```json
{"score": 73}
```
where "score" is an integer from 0 (certain to fail or score poorly) to 100 (certain to run and score among the best), using the full range so that different implementations get different scores.
\end{promptbox}

\begin{promptbox}[mblueL]{mblueD}{Tournament judge}{box:sr-tournament}
You are deciding which of TWO candidate IMPLEMENTATIONS is more worth spending an expensive evaluation on, for the SAME task. Each is real code that will be executed on the task's data and scored by the task's metric. Judge which implementation is more likely to (1) run without error and produce a valid output of the required form, and (2) score well on the task's metric.

TASK SPECIFICATION (data/inputs provided, required output, metric):
<task specification>

IMPLEMENTATION A:
```python
<code of candidate A>
```

IMPLEMENTATION B:
```python
<code of candidate B>
```

Read both implementations carefully. Consider: does it correctly consume the provided data/inputs; does it actually compute something aligned with the task's objective (not a trivial constant, a degenerate output, or leakage of the target values); is it likely to run without crashing; and does its approach plausibly beat the baseline. Judge the ACTUAL code, not the comments' claims. Ignore length/verbosity and which one is labelled A or B.

First THINK briefly, comparing the two implementations directly on the points above (a few sentences).
THEN give your FINAL ANSWER as one JSON object in a ```json code fence:
```json
{"winner": "A"}
```
where "winner" is exactly "A" or "B".
\end{promptbox}

\begin{promptbox}[mblueL]{mblueD}{Setwise judge}{box:sr-setwise}
You are screening candidate IMPLEMENTATIONS of the SAME task before an expensive evaluation step. Each is real code that will be executed on the task's data and scored by the task's metric. Some of them will be DISCARDED without evaluation. Pick the ones that are LEAST likely to (1) run without error and produce a valid output of the required form, and (2) score well on the task's metric.

TASK SPECIFICATION (data/inputs provided, required output, metric):
<task specification>

IMPLEMENTATION [1]:
```python
<code of candidate 1>
```
...
IMPLEMENTATION [N]:
```python
<code of candidate N>
```

Read every implementation carefully. Consider: does it correctly consume the provided data/inputs; does it actually compute something aligned with the task's objective (not a trivial constant, a degenerate output, or leakage of the target values); is it likely to run without crashing; and does its approach plausibly beat the baseline. Judge the ACTUAL code, not the comments' claims. Ignore length/verbosity and the order in which they are listed.

First THINK briefly (a few sentences). THEN give your FINAL ANSWER as one JSON object in a ```json code fence with labels:
```json
{"discard": [4, 9, 2]}
```
\end{promptbox}

\begin{promptbox}[mblueL]{mblueD}{Listwise judge}{box:sr-listwise}
You are ranking candidate IMPLEMENTATIONS of the SAME task by how worth it is to spend an expensive evaluation on each. Each is real code that will be executed on the task's data and scored by the task's metric. Rank them by how likely each is to (1) run without error and produce a valid output of the required form, and (2) score well on the task's metric.

TASK SPECIFICATION (data/inputs provided, required output, metric):
<task specification>

IMPLEMENTATION [1]:
```python
<code of candidate 1>
```
...
IMPLEMENTATION [N]:
```python
<code of candidate N>
```

Read every implementation carefully. Consider: does it correctly consume the provided data/inputs; does it actually compute something aligned with the task's objective (not a trivial constant, a degenerate output, or leakage of the target values); is it likely to run without crashing; and does its approach plausibly beat the baseline. Judge the ACTUAL code, not the comments' claims. Ignore length/verbosity and the order in which they are listed.

First THINK briefly (a few sentences). THEN give your FINAL ANSWER as one JSON object in a ```json code fence, listing ALL labels from BEST to WORST:
```json
{"ranking": [3, 1, 7, ...]}
```
\end{promptbox}

\section{Intuition on BVSR}

\subsection{Information-Induced Anchoring and Proposal Originality}
\label{app:bv_theory}

\begin{assumption}[Shared-Signal Generation]
\label{ass:shared_signal}
Let $b=\mathrm{info}(I)$ and $\mathcal{Z}=\{z_1,\ldots,z_K\}$ denote a fixed set of semantic idea classes.
Each idea $z_i$ has a task-association score $a_i$. For generation $t$, the model selects:
\begin{equation}
Z_t(b)
=
\arg\max_i
\left\{
\lambda(b)a_i+\varepsilon_{t,i}
\right\},
\end{equation}
where $\varepsilon_{t,i}$ are i.i.d.\ Gumbel random variables and $\lambda(b)\geq0$ is differentiable and non-decreasing, i.e., $\lambda'(b)\geq0$. Thus, additional task information increases the influence of the shared task signal relative to generation-specific exploratory variation.
\end{assumption}

Under Assumption~\ref{ass:shared_signal}, the induced probability of generating idea $z_i$ is
\begin{equation}
\pi_b(z_i)
=
\frac{\exp(\lambda(b)a_i)}
{\sum_{j=1}^{K}\exp(\lambda(b)a_j)}.
\label{eq:shared_signal_softmax}
\end{equation}
Thus, increasing task information strengthens the common task signal shared across generations relative to independent exploratory variation. 

\begin{proposition}
Under Assumption~\ref{ass:shared_signal}, the expected originality of generated proposals is non-increasing with task information:
\begin{equation}
\frac{d}{d\,\mathrm{info}(I)}
\mathbb{E}[o(c)]
\leq 0.
\end{equation}
Moreover, for a fixed generation budget $n$, when higher-utility ideas are less strongly associated with the shared task signal, increasing task information also decreases the expected utility of the best generated proposal.
\end{proposition}

\begin{proof}
The proof proceeds in three steps. We first derive the proposal distribution induced by shared-signal generation. We then show that stronger task information reduces expected originality. Finally, we characterize when the same concentration effect reduces the expected utility of the best proposal under a fixed generation budget.

\paragraph{Induced proposal distribution.}
Under Assumption~\ref{ass:shared_signal}, generation $t$ selects
$Z_t(b)=\arg\max_i\{\lambda(b)a_i+\varepsilon_{t,i}\}$.
By the Gumbel-max identity, the probability of selecting idea $z_i$ is
\begin{equation}
\pi_b(z_i)
=
\Pr\!\left(Z_t(b)=z_i\right)
=
\frac{\exp(\lambda(b)a_i)}
{\sum_{j=1}^{K}\exp(\lambda(b)a_j)},
\end{equation}
which means increasing $b$ strengthens the shared task signal relative to generation-specific variation.

\paragraph{Effect on originality.}
Consider $n>1$ independently generated proposals. Let $N_i$ denote the number of proposals belonging to semantic idea class $z_i$, with empirical frequency $\widehat{\pi}_i=N_i/n$. For any proposal $c$ in class $z_i$, we have $|g(c)|=N_i$, so its log-scaled originality is
$o(c)=1-\log N_i/\log n=-\log\widehat{\pi}_i/\log n$.
Averaging over all proposals gives
\begin{equation}
\frac{1}{n}\sum_c o(c)
=
-\frac{1}{\log n}
\sum_{i=1}^{K}\widehat{\pi}_i\log\widehat{\pi}_i
=
\frac{H(\widehat{\pi})}{\log n},
\label{eq:originality_entropy}
\end{equation}
where $H(\widehat{\pi})=-\sum_{i=1}^{K}\widehat{\pi}_i\log\widehat{\pi}_i$ is the empirical entropy of the semantic idea distribution. It therefore suffices to show that the expected empirical entropy is non-increasing in $b$.

Without loss of generality, order the task-association scores as
$a_1\geq a_2\geq\cdots\geq a_K$.
For any $r<K$, let
$S_r(\lambda)=\sum_{i=1}^{r}\pi_b(z_i)$
denote the probability assigned to the $r$ most task-associated ideas. Differentiating Eq.~\ref{eq:shared_signal_softmax} gives
\begin{equation}
\frac{dS_r}{d\lambda}
=
S_r(1-S_r)
\left(
\mathbb{E}[a\mid i\leq r]
-
\mathbb{E}[a\mid i>r]
\right)
\geq 0.
\label{eq:anchoring_concentration}
\end{equation}
The inequality holds since the first $r$ ideas have no smaller association scores than the rest. Hence, increasing $\lambda$ shifts probability mass toward the most task-associated ideas, so $\pi_b$ becomes increasingly concentrated in the sense of majorization.

We next show that this concentration cannot increase the expected empirical entropy. Since $(N_1,\ldots,N_K)$ follows a multinomial distribution with probabilities $\pi_b$, each marginal satisfies
$N_i\sim\mathrm{Binomial}(n,\pi_b(z_i))$.
Define
$h_n(p)=\mathbb{E}_{N\sim\mathrm{Binomial}(n,p)}
[-\frac{N}{n}\log\frac{N}{n}]$.
Then
\begin{equation}
\mathbb{E}[H(\widehat{\pi})]
=
\sum_{i=1}^{K}
h_n\!\left(\pi_b(z_i)\right).
\label{eq:expected_empirical_entropy}
\end{equation}
The function $h_n$ is the Bernstein polynomial of the concave function
$h(x)=-x\log x$, and is therefore concave. Consequently,
$\mathbb{E}[H(\widehat{\pi})]$ is symmetric and concave in $\pi_b$, and hence Schur-concave. Since increasing $\lambda$ makes $\pi_b$ more concentrated,
$\frac{d}{d\lambda}\mathbb{E}[H(\widehat{\pi})]\leq0$.
Together with $\lambda'(b)\geq0$, the chain rule gives
\begin{equation}
\frac{d}{db}\mathbb{E}[H(\widehat{\pi})]
=
\frac{d}{d\lambda}\mathbb{E}[H(\widehat{\pi})]\lambda'(b)
\leq0.
\end{equation}
Combining this with Eq.~\ref{eq:originality_entropy} and $\log n>0$ yields
\begin{equation}
\frac{d}{db}
\mathbb{E}\left[\frac{1}{n}\sum_c o(c)\right]
\leq0.
\label{eq:originality_information}
\end{equation}

\paragraph{Effect on the best proposal.}
We finally characterize how the same concentration affects discovery quality. Let each semantic idea class $z_i$ have utility $u_i$, and reorder the classes such that
$u_1\leq u_2\leq\cdots\leq u_K$.
Suppose higher-utility ideas are weakly less associated with the shared task signal, so that
$a_1\geq a_2\geq\cdots\geq a_K$.
Let
$F_j(\lambda)=\sum_{i=1}^{j}\pi_b(z_i)$
denote the probability of generating an idea with utility at most $u_j$.
By the same argument as Eq.~\ref{eq:anchoring_concentration},
\begin{equation}
\frac{dF_j}{d\lambda}\geq0
\qquad
\text{for all } j<K.
\label{eq:utility_cdf_concentration}
\end{equation}

Let $M_n(b)=\max_{t=1,\ldots,n}u(Z_t(b))$ denote the utility of the best idea among $n$ independent generations. Since
$\Pr(M_n(b)\leq u_j)=F_j(\lambda)^n$, its expectation can be written as
\begin{equation}
\mathbb{E}[M_n(b)]
=
u_K
-
\sum_{j=1}^{K-1}
(u_{j+1}-u_j)F_j(\lambda)^n.
\label{eq:best_of_n}
\end{equation}
Differentiating with respect to $\lambda$ gives
\begin{equation}
\frac{d}{d\lambda}\mathbb{E}[M_n(b)]
=
-
n\sum_{j=1}^{K-1}
(u_{j+1}-u_j)
F_j(\lambda)^{n-1}
\frac{dF_j}{d\lambda}
\leq0.
\end{equation}
Since $\lambda'(b)\geq0$, we therefore obtain
\begin{equation}
\frac{d}{db}\mathbb{E}[M_n(b)]
\leq0.
\end{equation}

Thus, stronger task information reduces expected proposal originality, and when higher-utility ideas are less aligned with the shared task signal, it also reduces the expected utility of the best proposal within a fixed generation budget.
\end{proof}

\subsection{Utility-Gap Advantage in Selective Retention}
\label{app:sr_utility_gap}

\begin{assumption}[Utility-Gap Comparison]
\label{ass:utility_gap}
For any pair of candidates with utilities $u(c_i)$ and $u(c_j)$, let
$\Delta=|u(c_i)-u(c_j)|$ denote their utility gap. We assume that the probability of correctly ordering the pair, $P_{\mathrm{correct}}(\Delta)$, is strictly increasing in $\Delta$. 

Thus, candidates with larger utility gaps are easier to distinguish.
\end{assumption}

\begin{proposition}
Let the $n$ candidates be ordered by utility as
$u(c_{(1)})\leq\cdots\leq u(c_{(n)})$.
For a discard ratio $d$, let $m=\lfloor dn\rfloor$ and define the utility gap across the selective-retention boundary as
$\Delta_{\mathrm{SR}}(d)=u(c_{(m+1)})-u(c_{(m)})$.
We further define the gap between the best candidate and its closest competitor as
$\Delta_{\mathrm{best}}=u(c_{(n)})-u(c_{(n-1)})$.
Under Assumption~\ref{ass:utility_gap}, if
$\Delta_{\mathrm{SR}}(d)>\Delta_{\mathrm{best}}$, then
\begin{equation}
P_{\mathrm{correct}}\!\left(\Delta_{\mathrm{SR}}(d)\right)
>
P_{\mathrm{correct}}\!\left(\Delta_{\mathrm{best}}\right).
\end{equation}
\end{proposition}

\begin{proof}
The proof follows directly from the utility-gap comparison assumption. Selective retention only needs to separate candidates across the discard boundary. At discard ratio $d$, the closest such pair is $c_{(m)}$ and $c_{(m+1)}$, whose utility gap is
$\Delta_{\mathrm{SR}}(d)=u(c_{(m+1)})-u(c_{(m)})$.
In contrast, identifying the best candidate requires distinguishing $c_{(n)}$ from its closest utility competitor $c_{(n-1)}$, with gap
$\Delta_{\mathrm{best}}=u(c_{(n)})-u(c_{(n-1)})$.

By Assumption~\ref{ass:utility_gap}, $P_{\mathrm{correct}}(\Delta)$ is strictly increasing in $\Delta$. Therefore, whenever
$\Delta_{\mathrm{SR}}(d)>\Delta_{\mathrm{best}}$, we have
\begin{equation}
P_{\mathrm{correct}}\!\left(\Delta_{\mathrm{SR}}(d)\right)
>
P_{\mathrm{correct}}\!\left(\Delta_{\mathrm{best}}\right).
\end{equation}
Thus, when the utility separation at the selective-retention boundary is larger than the separation among the top candidates, screening across the retention boundary is more accurate than identifying the single best candidate.
\end{proof}

\subsection{Safe Discarding under Selective Retention}
\label{app:sr_safe_discard}

\begin{proposition}
Let $d$ denote the overall discard ratio and $\beta_q(d)$ the expected fraction of top-$q$ proposals discarded by selective retention. For a tolerated loss $\epsilon$, define
\begin{equation}
d_q^\star(\epsilon)
=
\sup\{d:\beta_q(d)\leq\epsilon\}.
\label{eq:safe_discard_threshold}
\end{equation}
Then $d_q^\star(\epsilon)$ is the largest discard ratio that limits the expected loss of top-$q$ proposals to at most $\epsilon$, and therefore minimizes the number of proposals requiring verification under this constraint.
\end{proposition}

\begin{proof}
For a fixed ranking of $n$ proposals, let $\mathcal{D}(d)$ denote the set of proposals discarded at ratio $d$. Since selective retention discards proposals from the bottom of the same ranking, for any $d_1\leq d_2$,
$\mathcal{D}(d_1)\subseteq\mathcal{D}(d_2)$. Hence, the expected fraction of top-$q$ proposals discarded, $\beta_q(d)$, is non-decreasing in $d$.

For a tolerated loss $\epsilon$, define the feasible set
$\mathcal{F}_q(\epsilon)=\{d:\beta_q(d)\leq\epsilon\}$. By definition, $d_q^\star(\epsilon)$ is the largest feasible discard ratio.\footnote{For a finite set of candidates, discard ratios lie on a finite grid, so the supremum is attained.} Therefore,
\begin{equation}
\beta_q\!\left(d_q^\star(\epsilon)\right)
\leq \epsilon.
\end{equation}

At discard ratio $d$, $n(1-d)$ proposals remain for verification. Thus, choosing $d_q^\star(\epsilon)$ minimizes the number of proposals requiring verification while keeping the expected loss of top-$q$ proposals within $\epsilon$.
\end{proof}

\section{Prompt Templates}
\label{app:prompts}

\subsection{Blind Variation}
\label{app:prompts_bv}

Prompt~\ref{box:stageA} is the proposal prompt. \texttt{\{task\_context\}} is the task context. Every ideation turn also carries the literature-search tool.

\begin{promptbox}[mblueL]{mblueD}{Proposal Prompt}{box:stageA}
=== TOOLS ===
[
  {
    "type": "function",
    "function": {
      "name": "search_literature",
      "description": "Search the scientific literature (all disciplines) for real, published papers. Returns title, authors, year, citation count and abstract for each hit. Call it as many times as useful, refining your query between calls.",
      "parameters": {
        "type": "object",
        "properties": {
          "query": {
            "type": "string",
            "description": "search query (keywords or a phrase)"
          },
          "n": {
            "type": "integer",
            "description": "how many papers to return (default 6, max 15)"
          }
        },
        "required": [
          "query"
        ]
      }
    }
  }
]

=== USER MESSAGE ===
{task_context}Find ONE real paper from the field {field} whose single core mechanism is most promising to transplant into a different algorithm.

Use the search_literature tool to find real papers in this field. Search as many times as you find useful, refining your queries as you learn what exists.

Then reply EXACTLY in this format:
PAPER: <authors, title, year>
MECHANISM: <its ONE core mechanism, in one sentence>

ONE paper, ONE mechanism - do NOT combine several. Do NOT write any code.
\end{promptbox}

Prompt~\ref{box:stageB} is to implement the proposal. \texttt{\{task\_interface\}}, \texttt{\{baseline\_pool\}} and \texttt{\{task\_guidance\}} form the full task specification, and \texttt{\{output\_format\}} the code the task expects.

\begin{promptbox}[mblueL]{mblueD}{Implementation Prompt}{box:stageB}
{task_interface}

{baseline_pool}

# YOUR JOB 
PAPER: {paper}
MECHANISM: {mechanism}

Implement a REAL transplant of THAT ONE mechanism. ONE mechanism only; do NOT add other tricks.{task_guidance}

OUTPUT FORMAT (exactly):
```python
{output_format}
```
\end{promptbox}

\subsection{Selective Retention}
\label{app:prompts_sr}

The SR selector and the SFT regressor read the same input, Prompt~\ref{box:sr-input}, including \texttt{\{task\_specification\}} and the candidate's complete code. Both map this input to a single score.

\begin{promptbox}[mblueL]{mblueD}{Input of the SR selector and the SFT regressor}{box:sr-input}
### TASK
{task_specification}

### IMPLEMENTATION
{code}

### PREDICTED PERCENTILE OF VERIFIED SCORE:
\end{promptbox}

\section{Examples of Contexts under Different Information Levels}
\label{app:info-examples}

The information levels differ only in the task context at the start of the proposal generation message (Prompt~\ref{box:stageA}).
Prompts~\ref{box:info-ex2-l1}--\ref{box:info-ex1-l2} show example contexts for two tasks.

\begin{promptbox}[mblueL]{mblueD}{Ideation message, \textsc{SVG} (11.2 bits)}{box:info-ex2-l1}
Context: the samples sit on a 2-D lattice, and the goal is to rank features by how much of their variation is organized across the lattice rather than noise.
\end{promptbox}

\begin{promptbox}[mblueL]{mblueD}{Ideation message, \textsc{SVG} (22.2 bits)}{box:info-ex2-l2}
Context: this is spatially-variable-gene detection in spatial transcriptomics - ranking genes by whether their expression varies in a spatially structured way across a tissue section.
\end{promptbox}

\begin{promptbox}[mblueL]{mblueD}{Ideation message, \textsc{ZAPBench-16} (28.6 bits)}{box:info-ex1-l1}
CONTEXT: You are helping improve a procedure that reads a short stretch of many simultaneously recorded numeric signals and continues each of them forward for 16 further steps.
\end{promptbox}

\begin{promptbox}[mblueL]{mblueD}{Ideation message, \textsc{ZAPBench-16} (47.4 bits)}{box:info-ex1-l2}
CONTEXT: The procedure sees the last four time steps of tens of thousands of simultaneously recorded signals and must predict the next 16 steps for every one of them jointly. It is judged by mean absolute error against held-out continuations, averaged over windows, over the forecast steps and over all signals.
\end{promptbox}

\section{Human Validation Protocols}
\label{app:human_validation}

This section details the human evaluation protocols used in Section~\ref{sec:human_validation}.

\subsection{Originality}

Our originality measure depends on grouping proposals that express the same underlying scientific or algorithmic idea. To validate this semantic-equivalence step, we construct a set of 150 proposal pairs, balanced between pairs grouped together and pairs assigned to different groups by the LLM evaluator. The pairs are sampled across tasks, proposer models, and generation methods.

Five PhD-level researchers each evaluate 30 pairs. For each pair, the evaluator judges whether the two proposals express essentially the same underlying mechanism, ignoring differences in wording and implementation details. Pairs marked \textit{Unsure} (4 in total) are excluded from the agreement analysis. An example evaluation sheet is provided in Appendix~\ref{app:sheet_ori}.

\subsection{Non-obviousness}

We sample 150 high-quality proposals from BV and AR, defined as proposals whose verified score exceeds the best baseline of their task. We remove all method identifiers and ask five PhD-level researchers to evaluate their non-obviousness.

For each proposal, evaluators are shown the task name, a summary of the source paper, and a summary of the proposed method. Both summaries are rewritten by Claude Opus 5 into a uniform style to reduce cues that could reveal the generating method. Evaluators rate non-obviousness on a 1--5 scale, considering how readily the method's effectiveness on the task could be inferred from the source paper and their own knowledge of the literature. Higher scores indicate greater non-obviousness.

Proposals are split into forms of 30, each balanced across tasks and methods and presented in random order. Each evaluator completes one form. An example evaluation sheet is provided in Appendix~\ref{app:sheet}.

\section{Human Evaluation Example Form}

\subsection{Human Evaluation Example Form of Originality}
\label{app:sheet_ori}

\begin{figure}[h]
\centering
\includegraphics[width=\linewidth]{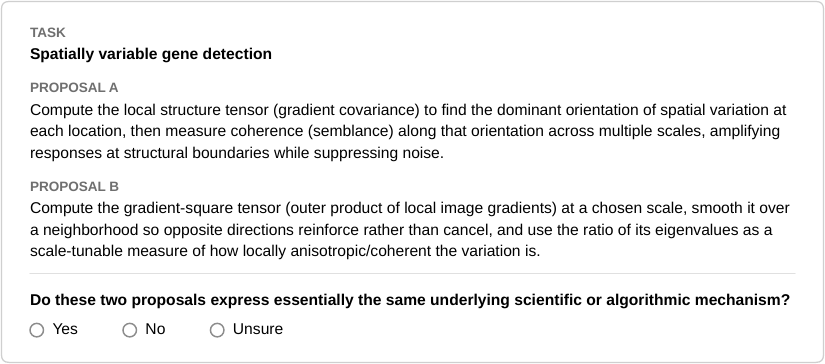}
\caption{\textbf{Example page of the originality evaluation form.} The evaluator sees a task and the mechanism descriptions of two proposals written for it, and judges whether the two express essentially the same underlying mechanism (Yes, No, or Unsure).}
\label{fig:sheet_ori}
\end{figure}

Each page gives the task name and the one-sentence mechanism descriptions of two proposals written for that task, followed by one multiple-choice question: do the two proposals express essentially the same underlying scientific or algorithmic mechanism (Yes / No / Unsure)? Both proposals of a pair come from the same task, the method that produced them and whether the grouping placed them together are not shown, and pairs appear in random order. Figure~\ref{fig:sheet_ori} shows the first pair page of one form.
  
\subsection{Human Evaluation Example Form of Non-Obviousness}
\label{app:sheet}

\begin{figure}[h]
\centering
\includegraphics[width=\linewidth]{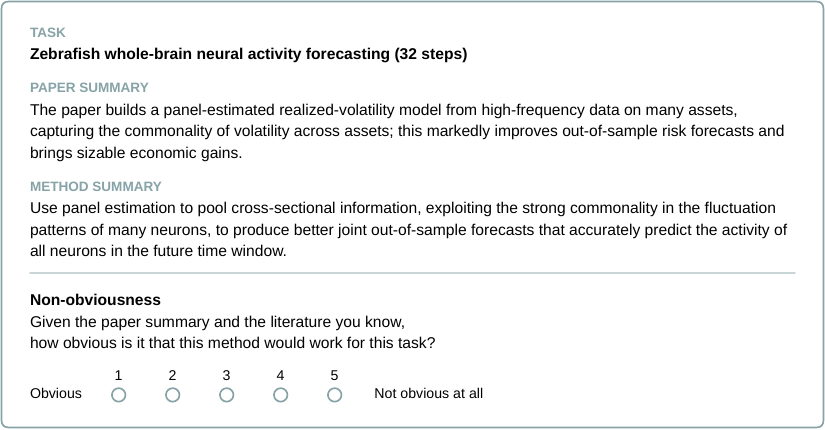}
\caption{\textbf{Example page of the non-obviousness evaluation form.} The evaluator sees a task, a summary of the paper the proposal draws on, and a summary of the proposed method, and rates how obvious it is that the method would work for the task, from 1 (obvious: the literature already shows it working on this kind of task) to 5 (not obvious at all).}
\label{fig:sheet}
\end{figure}

Evaluators rate proposals in a Google Form. The form opens with a short instruction page and one field for the evaluator's identifier; it then shows one proposal per page, 30 pages in total. Each page gives the task name, a summary of the paper the proposal draws on, and a summary of the proposed method, followed by a 1--5 question on non-obviousness. The method that produced each proposal is not shown, proposals appear in random order, and summaries are written by Claude Opus 5 in one uniform style so that the wording does not reveal the method. Figure~\ref{fig:sheet} shows the first page of one form.

\end{document}